\documentclass{article}

\PassOptionsToPackage{numbers}{natbib}
\usepackage[dblblindworkshop,final]{neurips_2026}
\workshoptitle{VLM4RWD}

\usepackage[utf8]{inputenc}
\usepackage[T1]{fontenc}
\usepackage{hyperref}
\usepackage{url}
\usepackage{booktabs}
\usepackage{amsfonts}
\usepackage{nicefrac}
\usepackage{microtype}
\usepackage{xcolor}
\usepackage{graphicx}
\usepackage{amsmath}
\usepackage{capt-of}
\usepackage{subcaption}
\usepackage{caption}
\usepackage{pifont}
\usepackage{multirow}

\usepackage[most]{tcolorbox}
\newtcolorbox{promptbox}{
    colback=gray!5,
    colframe=gray!40,
    boxrule=0.5pt,
    arc=2pt,
    left=6pt,
    right=6pt,
    top=6pt,
    bottom=6pt,
    fontupper=\small\ttfamily
}

\newcommand{\cmark}{\textcolor{green!55!black}{\ding{51}}}
\newcommand{\xmark}{\textcolor{red!75!black}{\ding{55}}}
\newcommand{\pmark}{\textcolor{orange!90!black}{$\sim$}}

\newcommand{\myparagraph}[1]{\vspace{3pt}\noindent{\bf #1}}

\title{OneScene-Bench: Do Vision-Language Models Integrate Evidence Across Views?}

\author{
  Swathi Suhas \quad
  Anurag Das \quad
  Bernt Schiele \quad
  Jonas Fischer\\ [2mm]
  Max Planck Institute for Informatics, \\
  Saarland Informatics Campus, Saarbrücken, Germany
  }

\begin{document}

\maketitle

\begin{abstract}
Vision-language models (VLMs) increasingly process multiple views of the same scene, but it remains unclear whether they integrate these observations into a coherent scene-level representation. Existing multi-image evaluations do not isolate this capability, as tasks may admit correct solutions through independent per-view processing without requiring cross-view correspondence. We introduce \textbf{OneScene-Bench}, a controlled benchmark designed to explicitly distinguish when cross-view integration is necessary. Our benchmark evaluates three complementary capabilities on real-world multi-view scenes: unique object counting, cross-view consistency, and spatial reasoning. For counting, we control object overlap across views to separate cases solvable by simple per-view aggregation from those requiring instance-level correspondence. For consistency, we introduce localized edits in shared and view-specific regions, testing whether models distinguish genuine scene contradictions from differences caused by viewpoint and visibility. For spatial reasoning, we construct object relations that require combining spatial evidence across viewpoints to compare physical proximity. We evaluate twelve state-of-the-art open-weight and proprietary VLMs and find that counting accuracy drops substantially on complementary views, ranging from 4.5--50.0\%, compared with 45.5--81.5\% on complete-overlap examples. Moreover, cross-view consistency remains around 50\% across models, including recent proprietary VLMs, despite substantially stronger performance on spatial reasoning. Our results reveal a substantial gap between processing multiple images and integrating evidence across them, highlighting the need for evaluations that explicitly test scene-level multi-view understanding. Code and data are available at \url{https://github.com/swathisuhas/OneScene-Bench}.

\end{abstract}
\section{Introduction}

Vision-language models~(VLMs) increasingly support multiple images as input, enabling them to reason over different observations of the same physical scene. Unlike generic multi-image understanding, multi-view scene understanding requires models to recognize repeated objects across viewpoints, account for viewpoint-dependent visibility, and combine complementary observations into a coherent representation of the scene. These capabilities are particularly important in embodied settings such as robotics and autonomous systems, where an agent observes the same environment from multiple viewpoints but must reason about it as a single scene.

However, multi-view input does not necessarily imply multi-view reasoning. Consider counting unique objects across two views. A model should recognize repeated instances across viewpoints and count each physical object only once. Yet when both views contain the same objects, or one contains only a subset of the other, taking the larger per-view count gives the correct answer. When each view contains instances not visible in the other, the model must instead determine which objects are shared to recover the total. A model with scene-level understanding should correctly integrate information across different configurations, regardless of how objects are distributed across the views. This reveals a fundamental ambiguity in multi-view evaluation: a correct answer does not necessarily show cross-view integration. Recent benchmarks evaluate reasoning across multiple images and viewpoints in diverse environments~\cite{chen2026triviewbench,lee2025spatialmosaic,xu2026citycube,zhou2025urbench}. However, these evaluations do not explicitly distinguish between predictions that require cross-view integration and those that can be solved through independent per-view processing. As a result, strong benchmark performance does not necessarily mean that a model has integrated and correctly linked information across views. A reliable evaluation must therefore control whether cross-view integration is actually necessary to obtain the correct answer.

\begin{figure}
    \centering
    \includegraphics[width=\linewidth]{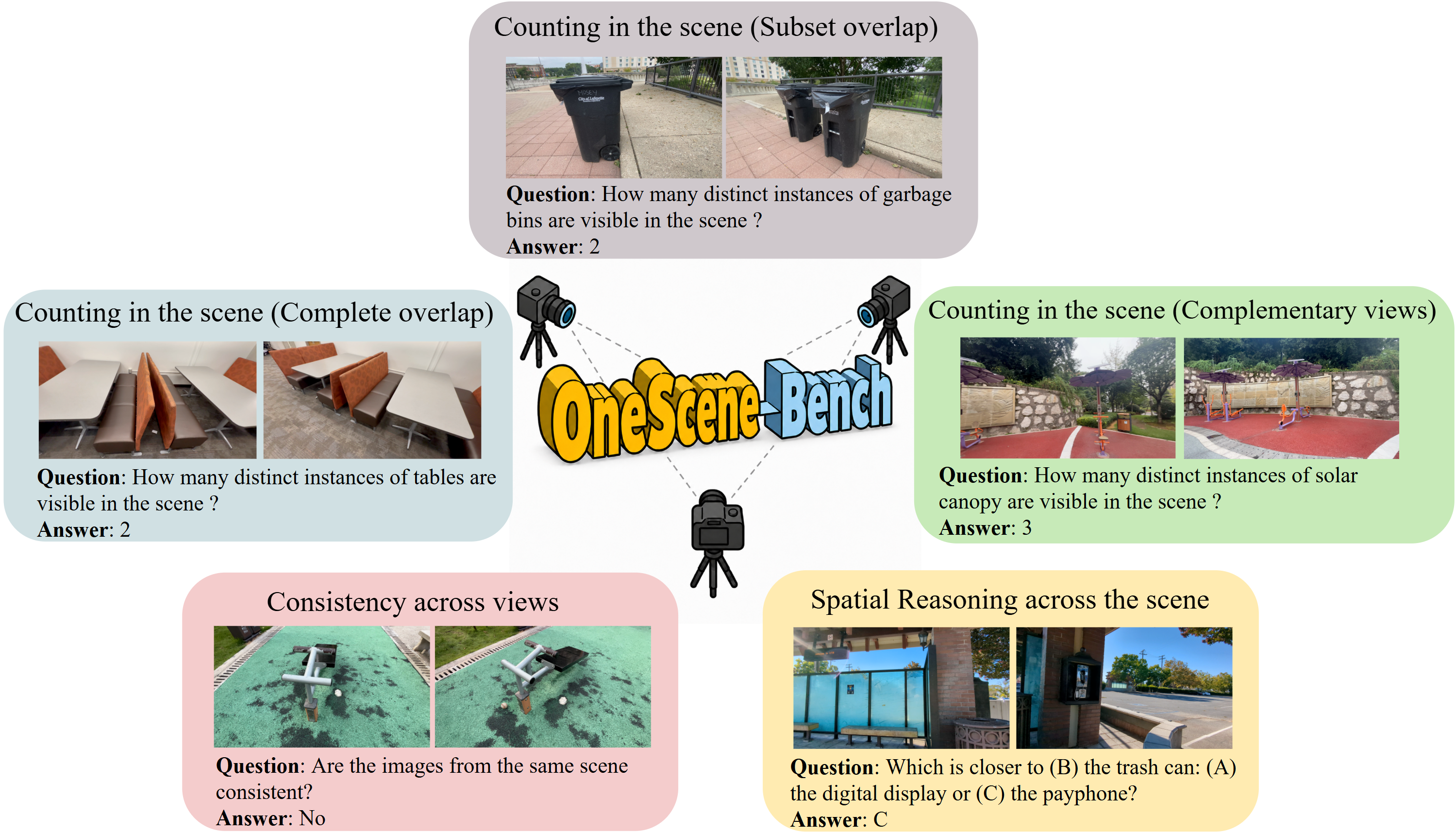}
    \caption{\textbf{Overview of OneScene-Bench.} OneScene-Bench evaluates whether VLMs correctly integrate and reason across views of the same physical scene through three tasks: unique object counting under complete-overlap, subset-overlap, and complementary-view configurations; cross-view consistency; and spatial reasoning. Representative image pairs, questions, and ground-truth answers are shown for each task.}
    \label{fig:teaser}
\end{figure}

We introduce \textbf{OneScene-Bench}, a human-annotated, controlled benchmark for evaluating whether VLMs reconcile multiple views as observations of the same physical scene~(Figure \ref{fig:teaser}). Our central design principle is to control the evidence required to solve each example. For multi-view counting, we construct three configurations of cross-view object overlap: \emph{complete overlap}, \emph{subset overlap}, and \emph{complementary views}. The first two admit a max-count shortcut, whereas in complementary views each image contributes instances not visible in the other, making per-view counts alone insufficient. Comparing performance across these configurations directly measures whether counting ability persists when cross-view integration becomes necessary. We probe the same capability with a second task, \textbf{cross-view consistency}. Differences between two views do not necessarily imply differences in the scene, as objects may be out of frame or occluded. We introduce localized edits either in regions visible from both viewpoints, creating a genuine contradiction between the views, or in regions visible from only one, leaving the pair compatible. This tests whether a model understands which evidence should agree across viewpoints, rather than merely detecting differences between images. Finally, \textbf{multi-view spatial reasoning} evaluates whether models can combine evidence across viewpoints to compare the physical proximity of objects. All examples are constructed from real-world multi-view captures~\cite{ling2024dl3dv} and manually verified.

We evaluate twelve open-weight and proprietary VLMs using an open-ended protocol and analyze both their performance and failure modes. Across all models, counting performance drops on complementary views: accuracy ranges from 45.5--81.5\% on complete overlap and 26.7--76.7\% on subset overlap, but only 4.5--50.0\% on complementary views, where cross-view correspondence and deduplication are required. Cross-view consistency accuracy remains around 50\%, i.e., random chance, across models, with models responding similarly to edits in shared and view-specific regions, indicating limited sensitivity to cross-view visibility. We further attribute counting predictions to simple per-view aggregation strategies and study sensitivity to image order, removal of visual input, and explicit reasoning. Together, these results reveal a clear gap between processing multiple images and correctly performing visual reasoning across them.

We summarize our contributions as follows:
\begin{itemize}
    \item We introduce \textbf{OneScene-Bench}, comprising 1,300 manually verified examples from real-world multi-view scenes across counting, consistency, and spatial reasoning.
    \item We develop \textbf{controlled diagnostics for cross-view integration}: an overlap-based counting taxonomy that distinguishes cases where per-view aggregation is sufficient from those where cross-view integration is necessary, and a consistency intervention that controls whether visual differences occur in regions shared across viewpoints.
    \item We provide a broad evaluation of \textbf{twelve VLMs} under an open-ended protocol, together with analyses of aggregation behavior, image-order sensitivity, no-image bias, explicit reasoning, and scaling with additional views, revealing a persistent gap between processing multiple images and integrating them into a coherent scene-level representation.
\end{itemize}

\section{Related Work}

\myparagraph{Multimodal Large Language Models (MLLMs).}
Early MLLMs established the main paradigms for integrating vision with large language models. Flamingo~\cite{alayrac2022flamingo} introduced gated cross-attention to condition an LLM on interleaved images and text, enabling multimodal few-shot learning. BLIP-2~\cite{li2023blip} proposed the Q-Former as an efficient interface between frozen vision encoders and LLMs, while LLaVA~\cite{liu2023visual} demonstrated the effectiveness of visual instruction tuning for general visual conversation and reasoning. Building on these foundations, proprietary models such as GPT-4V~\cite{openai2023gpt4v}, Gemini~\cite{team2023gemini,team2024gemini}, and Claude 3~\cite{anthropic2024claude3} demonstrated broad multimodal reasoning across images, videos, and other modalities. Similarly, open-weight models, including the Qwen-VL series~\cite{bai2023qwen,wang2024qwen2,bai2025qwen3}, LLaVA-OneVision~\cite{li2024llava}, InternVL~\cite{wang2025internvl3}, MiniCPM-V~\cite{yu2026minicpm}, GLM-4.6V~\cite{hong2025glm}, Phi-4-Multimodal~\cite{abouelenin2025phi}, and Kimi-VL~\cite{team2025kimi}, extended visual instruction tuning to high-resolution images, multiple images, videos, and long multimodal contexts. We evaluate representative models from both proprietary and open-weight families.

\myparagraph{Multi-Image Understanding Benchmarks.} Multi-image evaluation has progressed from task-specific benchmarks to broader evaluation suites. BLINK~\cite{fu2024blink}, Mementos~\cite{wang2024mementos}, and MileBench~\cite{dingjie2024milebench} focus on visual perception, image sequences, and long multimodal contexts, respectively. MIRB~\cite{zhao2024benchmarking}, MIBench~\cite{liu2024mibench}, MMIU~\cite{meng2025mmiu}, MIMIC~\cite{das2026more}, MUIRBench~\cite{wang2025muirbench}, and MMRB~\cite{cheng2025evaluating} further evaluate diverse perceptual, semantic, and reasoning relations across images. However, these benchmarks often combine related or sequential images rather than different views of the same physical scene. Multi-view benchmarks study this more specific setting across several domains. MV-RoboBench~\cite{feng2025seeing} targets spatial understanding and robotic execution, while UrBench~\cite{zhou2025urbench} and CityCube~\cite{xu2026citycube} evaluate urban understanding across ground-level, vehicle, aerial, and satellite views. All-Angles Bench~\cite{yeh2026seeing} includes counting and geometric reasoning in real-world scenes, whereas MultiView-Bench~\cite{zhang2026multiview} and TriViewBench~\cite{chen2026triviewbench} study world-centric reasoning and scene recovery in synthetic 3D environments. SpatialMosaic-Bench~\cite{lee2025spatialmosaic} focuses on integrating fragmented spatial evidence under occlusion and limited overlap. Although these benchmarks cover important aspects of multi-view understanding, they generally do not explicitly separate questions solvable through independent per-view processing from those requiring cross-view integration. OneScene-Bench isolates this distinction by controlling object overlap for counting, the visibility of localized edits for consistency, and the distribution of relational evidence for spatial reasoning. These controlled tasks test whether models reconcile multiple observations into a coherent object-centric representation of the same physical scene.

\section{OneScene-Bench}
\label{sec:onescene_bench}

We introduce \textbf{OneScene-Bench}, a controlled benchmark for evaluating whether vision-language models (VLMs) integrate multiple views into a coherent representation of the same physical scene. Our central design principle is to control whether cross-view integration is necessary to answer a question correctly. This distinction is important because some multi-view questions can be solved by processing each image independently and applying a simple aggregation rule, without establishing correspondence between observations. OneScene-Bench evaluates three complementary capabilities: \textbf{multi-view counting}, \textbf{cross-view consistency}, and \textbf{multi-view spatial reasoning}.

\begin{figure*}[t]
\centering

\begin{subfigure}[t]{0.49\textwidth}
\vspace{0pt}
\centering

\label{fig:benchmark_comparison}

\scriptsize
\setlength{\tabcolsep}{2.5pt}
\renewcommand{\arraystretch}{1.15}

\resizebox{\linewidth}{!}{%
\begin{tabular}{@{}lccccc@{}}
\toprule
\textbf{Benchmark}
& \shortstack{\textbf{Multi-}\\\textbf{view}}
& \shortstack{\textbf{Free-form}\\\textbf{responses}}
& \textbf{Diagnostic}
& \shortstack{\textbf{Cross-view}\\\textbf{consistency}}
& \shortstack{\textbf{Counting}\\\textbf{types}}\\
\midrule

\multicolumn{6}{@{}l}{\textit{Multi-image benchmarks}} \\
\addlinespace[2pt]

BLINK
& \pmark & \xmark & \pmark & \xmark & \xmark \\

MMIU
& \pmark & \xmark & \pmark & \xmark & \xmark \\

MIRB
& \xmark & \pmark & \pmark & \xmark & \xmark \\
\midrule

\multicolumn{6}{@{}l}{\textit{Multi-view benchmarks}} \\
\addlinespace[2pt]

All-Angles Bench
& \cmark & \xmark & \cmark & \pmark & \pmark \\

MultiView-Bench
& \cmark & \cmark & \cmark & \xmark & \xmark \\

SpatialMosaic-Bench
& \cmark & \pmark & \cmark & \xmark & \pmark \\

\textbf{OneScene-Bench}
& \cmark & \cmark & \cmark & \cmark & \cmark \\

\bottomrule
\end{tabular}%
}
\caption{\textbf{Benchmark Properties.} Comparison of representative multi-image and multi-view benchmarks with
OneScene-Bench across key benchmark characteristics. }

\end{subfigure}\hfill%
\begin{subfigure}[t]{0.49\textwidth}
\vspace{0pt}
\centering

\includegraphics[
    width=0.90\linewidth
]{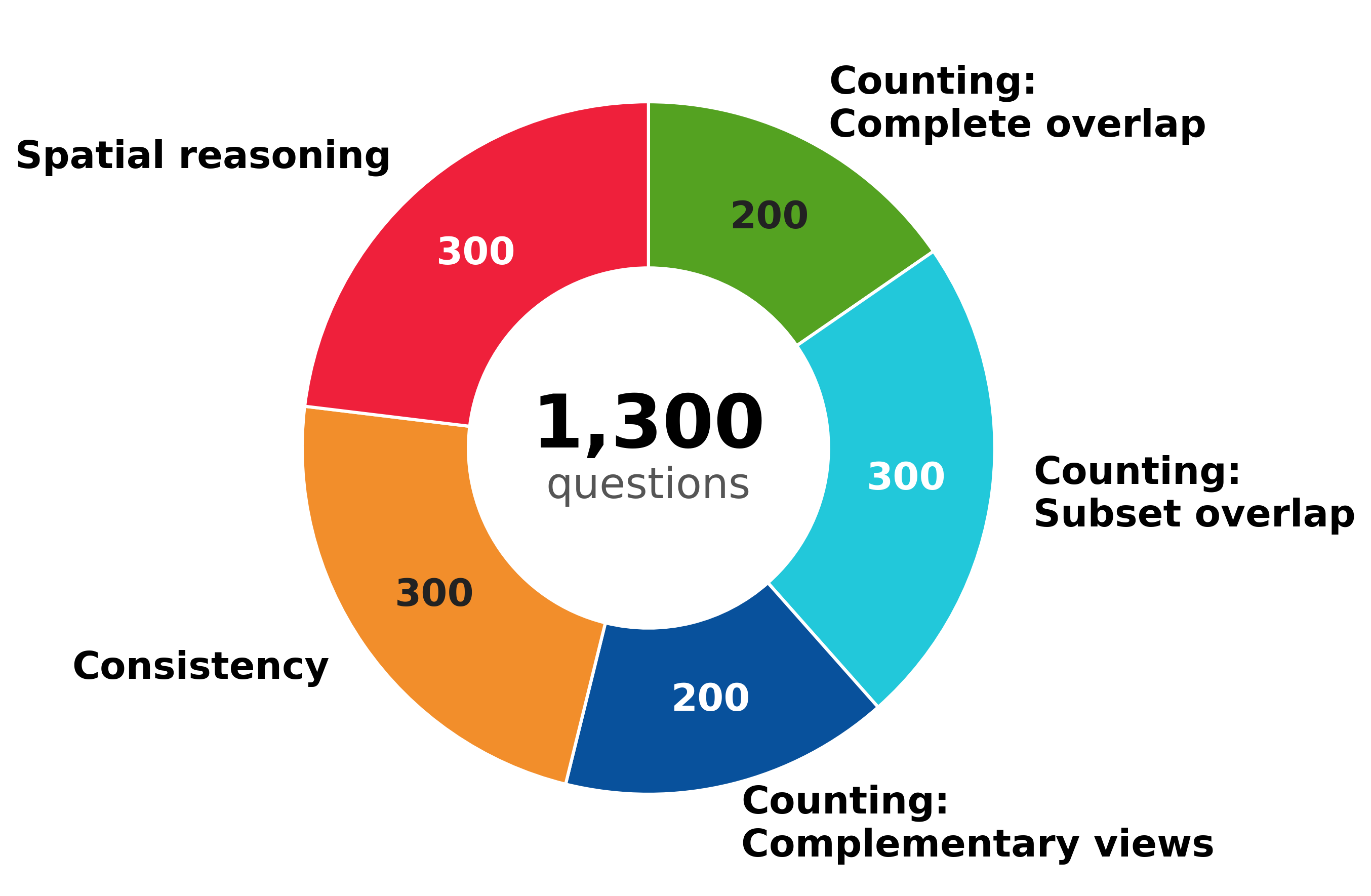}

\caption{\textbf{Benchmark composition.}}
\label{fig:composition}

\end{subfigure}

\caption{
\textbf{Properties and Composition of OneScene-Bench.}
\textbf{(a)} Comparison with representative multi-image and multi-view
benchmarks. \textit{Multi-view} denotes multiple observations of the same
physical scene; \textit{Free-form} denotes unrestricted responses;
\textit{Diagnostic} denotes controlled behavioral analyses; and
\textit{Cross-view consistency} denotes explicit consistency evaluation.
\textit{Counting types} denotes different counting scenarios covering different levels of view overlap.
Green ticks, orange tildes, and red crosses indicate full, partial, and no
coverage, respectively.
\textbf{(b)} Composition of the 1,300-question core two-view benchmark.
Multi-view counting contains 700 questions: 200 complete-overlap, 300
subset-overlap, and 200 complementary-view examples. Cross-view consistency
and multi-view spatial reasoning contain 300 questions each.
}
\label{fig:benchmark_overview}

\end{figure*}

The benchmark is constructed from DL3DV-10K~\cite{ling2024dl3dv}, which contains real-world scenes captured from multiple viewpoints. The core benchmark comprises \textbf{1,300 human-verified questions}, with two views per example: 700 counting questions, 300 consistency questions, and 300 spatial-reasoning questions (Fig.~\ref{fig:composition}). The required outputs are an integer count, a binary \textsc{yes}/\textsc{no} judgment, and a choice between two reference objects, respectively. We use model-assisted candidate mining (see Appendix~\ref{app:dataset_construction}) to identify suitable examples at scale, but all questions follow fixed human-defined templates and all annotations are manually verified.

\subsection{Multi-View Counting}
\label{sec:counting}

Given two views of the same scene and a target object category, the model must predict the number of \emph{unique physical instances} visible across the views. Let $S_1$ and $S_2$ denote the sets of target instances visible in each view. The scene-level count is
\begin{equation}
    C = |S_1 \cup S_2|.
    \label{eq:count}
\end{equation}
We control the relationship between $S_1$ and $S_2$ to distinguish cases that admit a simple per-view shortcut from those that require cross-view correspondence.

\paragraph{Complete overlap.}
Both views contain the same target instances, $S_1 = S_2$. The correct scene-level count equals either per-view count. We include 200 examples.

\paragraph{Subset overlap.}
One view contains a subset of the instances in the other, $S_1 \subset S_2$ or $S_2 \subset S_1$. The correct answer can still be recovered as the maximum of the two per-view counts. We include 300 examples.

\paragraph{Complementary views.}
Each view contains at least one instance absent from the other,
$S_1 \setminus S_2 \neq \emptyset$ and
$S_2 \setminus S_1 \neq \emptyset$.
Neither view alone contains all instances, and the maximum per-view count is insufficient. Correct prediction therefore requires identifying repeated instances across views while incorporating instances visible from only one viewpoint. We include 200 examples.

\paragraph{Solvability.}
All three configurations require aggregating observations across views to recover a scene-level count, but they differ in the cross-view information required for this aggregation. In complete overlap, both views contain the same set of physical instances, while in subset overlap, the instances in one view are contained in the other. In both cases, the union in Eq.~\eqref{eq:count} equals the larger per-view set, so the scene-level count can be recovered without explicitly estimating instance overlap. In complementary views, neither view contains the complete set of instances; recovering the union additionally requires identifying which observations correspond to the same physical instances across views, i.e., determining $|S_1 \cap S_2|$. These configurations therefore preserve the need for multi-view aggregation while controlling whether explicit instance-level correspondence is required.

\subsection{Cross-View Consistency}
\label{sec:consistency}

Given two views, the model determines whether they are compatible as observations of the \emph{same underlying scene}. This task requires reasoning about viewpoint-dependent visibility: an object appearing in only one image does not necessarily indicate a contradiction because its location may be outside the field of view or occluded in the other image. We construct controlled examples by introducing a localized object or visual artifact into one view using gpt-image-1.5~\cite{openai2025gptimage15}. We vary only whether the edited location should be observable from the other viewpoint. An edit in a \emph{view-specific region}, which is not observable from the other view, leaves the pair compatible with the same scene; the pair is labeled \textsc{consistent}. An edit in a \emph{shared observable region} creates a contradiction between the two observations; the pair is labeled \textsc{inconsistent}. Both conditions therefore contain an image-level difference, while differing in whether that difference implies a scene-level contradiction; detecting differences between images alone cannot separate them. We construct 300 balanced examples using this procedure.

\subsection{Multi-View Spatial Reasoning}
\label{sec:multiview_spatial}

The spatial-reasoning task evaluates whether VLMs can integrate spatial evidence across viewpoints. Each question specifies three objects, $A$, $B$, and $C$, and asks whether $B$ is physically closer to $A$ or to $C$. Notably, these objects might be spread over both views, for example, with $A$ and $B$ visible in one image and $B$ and $C$ in the other.
We construct 300 questions from pairs of views of the same DL3DV-10K~\cite{ling2024dl3dv} scene. Qwen3-VL-32B-Instruct~\cite{bai2025qwen3} proposes candidate object triplets for which the two views provide complementary evidence about relative proximity. We favor examples in which combining evidence across viewpoints informs the comparison beyond what is readily available from either view alone. Selected triplets are instantiated using a fixed question template and manually verified.

\subsection{Multi-View Scaling}
\label{sec:scaling}

The core benchmark uses two views per example. To study how counting behaves as observations accumulate, we extend the 200 complete-overlap questions to $N \in \{2, \dots, 6\}$ input views, keeping all target instances visible in every selected view so that the scene-level count is unchanged. Each additional view contributes another observation of the same physical instances while increasing the number of cross-view correspondences the model must resolve. This extension therefore measures whether additional viewpoints improve object recognition and scene understanding or instead increase identity confusion and double-counting. Evaluating every question in all five settings yields $200 \times 5 = 1{,}000$ view-conditioned instances; the two-view versions already belong to the core benchmark, so the extension adds 800 configurations, for 2,100 evaluation configurations in total.

\paragraph{Human Verification.}
\label{pg:hv}
All automatically proposed or generated examples undergo manual review before inclusion in OneScene-Bench; the VLM and image-generation model serve only to accelerate candidate discovery and controlled editing, and their outputs are never treated directly as ground truth. For counting, annotators verify the target category, the number of visible instances in each image, the identity of repeated instances across views, the overlap type, and the unique scene-level count. For consistency, annotators verify whether the edited region is shared between the viewpoints or visible from only one view, and whether the resulting consistency label is correct. For spatial reasoning, annotators confirm the identities of the three objects and the relative-proximity answer. Incorrect annotations are corrected, while examples with uncertain object identity, visibility, correspondence, or spatial relationships are removed.

\section{Evaluation}
\label{sec:evaluation}

\subsection{Evaluation Setup}

We evaluate all models in five settings in OneScene-Bench: complete-overlap counting, subset-overlap counting, complementary-view counting, cross-view consistency, and multi-view spatial reasoning. All models receive the same task instructions and input images, and we retain their complete generated responses.

We use GPT-4o~\cite{hurst2024gpt} as an LLM judge for all benchmark results. The judge receives the question, ground-truth answer, and complete model response, and determines whether the response reaches the correct conclusion. We report judge-based accuracy, allowing semantically equivalent answers and minor differences in response format. The exact prompts used are available in Appendix~\ref{app:prompts}.

\subsection{Benchmark Evaluation Results}

Table~\ref{tab:evaluation_results} reports accuracy across the five evaluation settings. Qwen3-VL-32B-Instruct~\cite{bai2025qwen3} drops from 70.0\% and 70.7\% on complete- and subset-overlap examples to 34.0\% on complementary views, while InternVL3.5-38B-HF~\cite{wang2025internvl3} drops from 72.0\% to 14.5\%. Even the strongest models exhibit this pattern: Gemini 3.7 Flash~\cite{gemini37flash2026} decreases from 81.5\% to 50.0\%, while GPT-5.5~\cite{openai2026gpt55} drops from 70.0\% to 38.5\%. The consistent degradation on complementary views identifies cross-view correspondence and deduplication as a key bottleneck, where models perform substantially better when the views contain largely redundant evidence than when information must be reconciled across views.

Consistency accuracy remains close to chance, ranging from 47.0\% to 55.7\%. LLaVA-OneVision-2-8B-Instruct~\cite{an2026llava} and Phi-4-multimodal-instruct~\cite{abouelenin2025phi} predict \textsc{yes} for all examples and consequently obtain 50.0\% on the balanced dataset. This behavior suggests that models often default to treating different views as consistent rather than verifying whether scene content agrees across viewpoints.

Spatial reasoning capabilities are substantially stronger for most models. GPT-5.5~\cite{openai2026gpt55} achieves the highest accuracy at 92.7\%, followed closely by Gemini 3.7 Flash~\cite{gemini37flash2026} at 89.7\%. Kimi-VL-A3B-Instruct is a notable outlier at 15.7\%, frequently selecting object $B$ rather than one of the required reference objects $A$ or $C$, indicating a failure to follow the specified answer format. Together, these results show that current VLMs can perform well on some multi-view tasks, yet struggle when they must consistently reconcile information across viewpoints.

\begin{table*}[t]
\centering
\small
\setlength{\tabcolsep}{1pt}

\begin{tabular*}{\textwidth}{
    @{\extracolsep{\fill}}llccccc@{}
}
\toprule
\textbf{Model}
&
& \multicolumn{3}{c}{\textbf{Counting} $\uparrow$}
& \textbf{Consistency $\uparrow$}
& \textbf{Spatial} \\
\cmidrule(lr){3-5}

&& \textbf{Complete}
& \textbf{Subset}
& \textbf{Complementary}
& & \textbf{Reasoning} $\uparrow$\\
\midrule

\parbox[t]{2mm}{\multirow{8}{*}{\rotatebox[origin=c]{90}{Open}}}
& Qwen3-VL-32B-Instruct~\cite{bai2025qwen3}
& 70.0 & \textbf{70.7} & 34.0 & 49.3 & 85.3 \\

&Gemma-4-31B-IT~\cite{team2026gemma}
& 61.0 & 63.3 & \textbf{35.0} & 50.3 & \textbf{86.3} \\

&InternVL3.5-38B-HF~\cite{wang2025internvl3}
& \textbf{72.0} & 57.3 & 14.5 & 50.7 & 80.7 \\

&LLaVA-OneVision-2-8B-Instruct~\cite{an2026llava}
& 55.5 & 53.0 & 18.0 & 50.0 & 77.0 \\

&MiniCPM-V-4.5~\cite{yu2026minicpm}
& 67.0 & 51.7 & 20.5 & 53.3 & 72.7 \\

&Phi-4-multimodal-instruct~\cite{abouelenin2025phi}
& 54.0 & 26.7 & 4.5 & 50.0 & 64.7 \\

&GLM-4.6V-Flash~\cite{hong2025glm}  & 57.0 & 56.7 & 28.0 & \textbf{55.7} & 81.7 \\

&Kimi-VL-A3B-Instruct~\cite{team2025kimi}
& 66.0 & 46.3 & 10.0 & 50.0 & 15.7 \\

\midrule

\parbox[t]{2mm}{\multirow{4}{*}{\rotatebox[origin=c]{90}{Closed}}}
&GPT-4o~\cite{hurst2024gpt}
& 61.5 & 52.3 & 28.5 & 47.0 & 73.0 \\

&GPT-5.5~\cite{openai2026gpt55}
& 70.0 & 69.3 & 38.5 & 54.3 & \textbf{92.7} \\

&Gemini 3.7 Flash~\cite{gemini37flash2026}
& \textbf{81.5} & \textbf{76.7} & \textbf{50.0} & 49.0 & 89.7 \\

&Claude Haiku 4.5~\cite{claudehaiku45_2025}
& 45.5 & 43.3 & 18.5 & \textbf{55.3} & 75.0 \\

\bottomrule
\end{tabular*}

\caption{
\textbf{VLMs on OneScene-Bench.}
Accuracy (\%) of twelve open-weight and proprietary VLMs on OneScene-Bench. The best-performing model in each group is shown in bold for every task. Across models, counting performance drops substantially on complementary views, where object correspondence and deduplication across viewpoints are required. Cross-view consistency remains around 50\%, whereas spatial reasoning is substantially stronger, revealing a marked difference in model capability across forms of multi-view integration.
}

\label{tab:evaluation_results}
\end{table*}

\subsection{Diagnostic Analyses}

We conduct additional analyses to examine whether model behavior is affected by input ordering, reliance on non-visual priors, explicit reasoning, simple per-view aggregation, and the number of input views.

\paragraph{No-Image Bias.}
We remove all input images while keeping the questions and evaluation procedure unchanged. Although accuracy generally drops, models retain non-trivial performance on several tasks, revealing substantial language and answer-distribution priors. Notably, LLaVA-OneVision-2 obtains 50\% consistency accuracy both with and without images by predicting \textsc{yes} for every example, showing that unchanged performance need not indicate successful use of visual evidence. Full results are reported in Table~\ref{tab:no_image_bias}; Appendix~\ref{app:refusal} lists how often models refuse to answer because no image is provided.

\begin{table}[t]
\centering
\captionsetup{position=bottom,skip=7pt}
\resizebox{\columnwidth}{!}{
\begin{tabular}{lccccc}
\toprule
& \multicolumn{3}{c}{\textbf{Counting}}
& \textbf{Consistency $\uparrow$}
& \textbf{Spatial} \\
\cmidrule(lr){2-4}
\textbf{Model}
& \textbf{Complete $\uparrow$}
& \textbf{Subset $\uparrow$}
& \textbf{Complementary $\uparrow$}
&
& \textbf{Reasoning $\uparrow$}\\
\midrule
Qwen3-VL-32B-Instruct~\cite{bai2025qwen3}
& 37.5 \textcolor{gray}{(70.0)}
& 26.7 \textcolor{gray}{(70.7)}
& 7.0 \textcolor{gray}{(34.0)}
& 50.0 \textcolor{gray}{(49.3)}
& 63.7 \textcolor{gray}{(85.3)} \\
Gemma-4-31B-IT~\cite{team2026gemma}
& 34.0 \textcolor{gray}{(61.0)}
& 28.3 \textcolor{gray}{(63.3)}
& 13.5 \textcolor{gray}{(35.0)}
& 0.0 \textcolor{gray}{(50.3)}
& 1.7 \textcolor{gray}{(86.3)} \\
InternVL3.5-38B-HF~\cite{wang2025internvl3}
& 41.0 \textcolor{gray}{(72.0)}
& 14.3 \textcolor{gray}{(57.3)}
& 1.5 \textcolor{gray}{(14.5)}
& 50.0 \textcolor{gray}{(50.7)}
& 55.3 \textcolor{gray}{(80.7)} \\
LLaVA-OneVision-2-8B-Instruct~\cite{an2026llava}
& 40.0 \textcolor{gray}{(55.5)}
& 17.7 \textcolor{gray}{(53.0)}
& 3.5 \textcolor{gray}{(18.0)}
& 50.0 \textcolor{gray}{(50.0)}
& 54.7 \textcolor{gray}{(77.0)} \\
\bottomrule
\end{tabular}
}
\caption{
\textbf{No-image bias.}
Accuracy (\%) when models receive the task instructions but no images; performance with images is shown in gray in parentheses. Models achieve non-trivial accuracy without visual input, indicating response biases and task priors, while visual evidence substantially improves counting and spatial reasoning performance.
}
\label{tab:no_image_bias}
\end{table}

\begin{figure*}[t]
\centering
\begin{subfigure}[t]{0.60\textwidth}
\vspace{0pt}
\centering

\resizebox{\linewidth}{!}{%
\begin{tabular}{lccc}
\toprule
& \multicolumn{3}{c}{\textbf{Counting Accuracy}} \\
\cmidrule(lr){2-4}
\textbf{Model}
& \textbf{Complete $\uparrow$}
& \textbf{Subset $\uparrow$}
& \textbf{Complementary $\uparrow$} \\
\midrule
Qwen3-VL-32B-Instruct~\cite{bai2025qwen3}          & \textbf{70.0} & \textbf{70.7} & \textbf{34.0} \\
Qwen3-VL-32B-Thinking~\cite{bai2025qwen3}
& 59.0 & 57.0 & 26.5 \\[2pt]
\cmidrule(lr){1-4}
Kimi-VL-A3B-Instruct~\cite{team2025kimi}
& \textbf{66.0} & \textbf{46.3} & 10.0 \\
Kimi-VL-A3B-Thinking~\cite{team2025kimi}
& 21.5 & 38.0 & \textbf{11.5} \\[2pt]
\cmidrule(lr){1-4}
MiniCPM-V-4.5~\cite{yu2026minicpm}
& \textbf{67.0} & \textbf{51.7} & \textbf{20.5} \\
MiniCPM-V-4.5 (Thinking)~\cite{yu2026minicpm}
& 61.5 & 44.3 & 9.5 \\
\bottomrule
\end{tabular}%
}

\caption{Reasoning-enabled variants.}
\label{fig:reasoning_ablation}
\end{subfigure}%
\hfill%
\begin{subfigure}[t]{0.39\textwidth}
\vspace{0pt}
\centering

\includegraphics[
    width=\linewidth
]{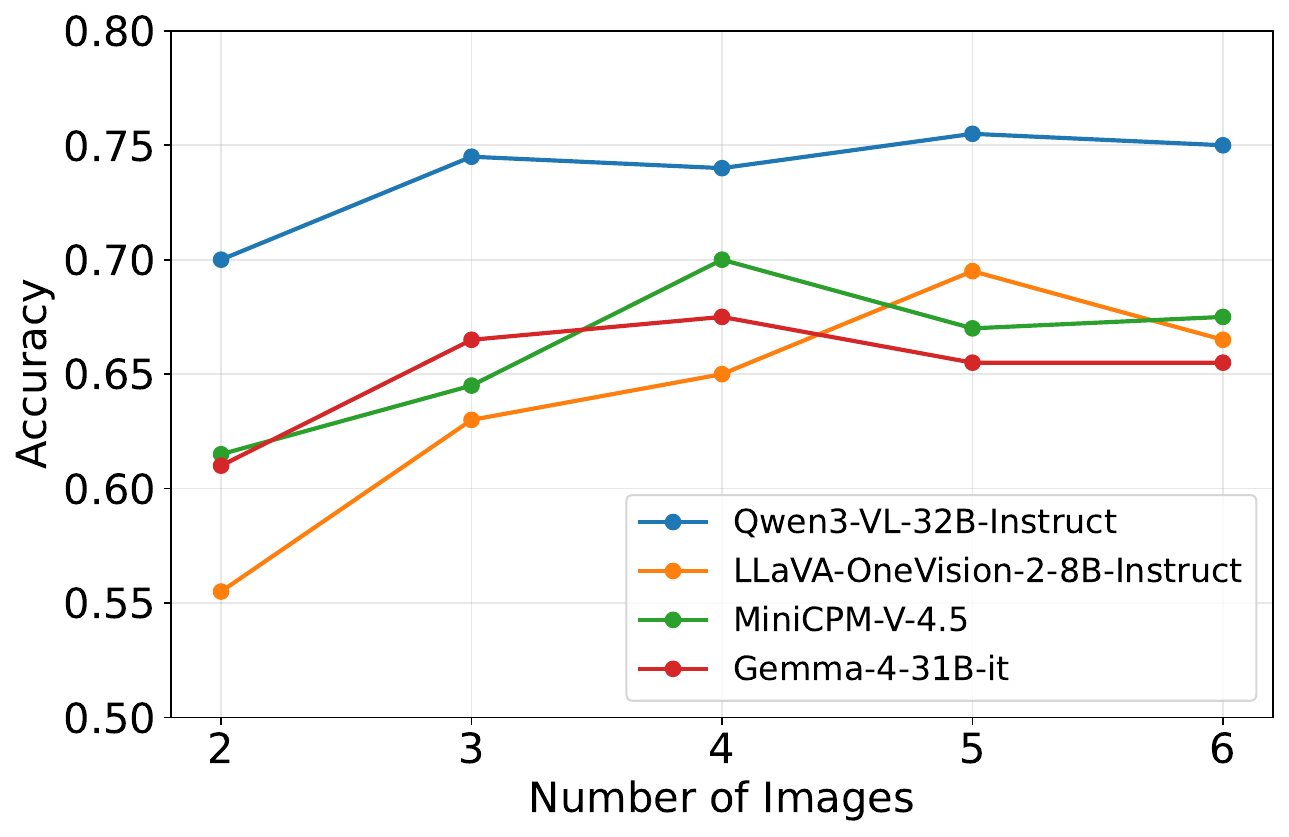}

\caption{Number of input views.}
\label{fig:scaling}
\end{subfigure}

\caption{
\textbf{Diagnostic analyses of reasoning and number of views.}
\textbf{(a)} Counting accuracy of instruction-tuned and reasoning-enabled variants. Reasoning generally reduces performance, indicating that explicit reasoning does not improve multi-view counting.
\textbf{(b)} Complete-overlap counting accuracy with $N=2$--$6$ input views. Performance initially improves with additional views but peaks before $N=6$ for all models, indicating that more views do not yield monotonic gains.
 }
\label{fig:counting_ablations}

\end{figure*}

\paragraph{Instruct vs.\ Thinking Variants.}
We compare matched reasoning-enabled and instruction-tuned variants of Qwen3-VL~\cite{bai2025qwen3} and Kimi-VL~\cite{team2025kimi}, together with direct and thinking-enabled inference for MiniCPM-V-4.5~\cite{yu2026minicpm}. As shown in Fig.~\ref{fig:reasoning_ablation}, reasoning reduces counting accuracy across nearly all models and settings; the only exception is a 1.5-point gain for Kimi-VL on complementary views. Thus, explicit reasoning does not systematically improve multi-view counting.

\paragraph{Comparison with Aggregation Heuristics.}

We test whether joint predictions can be explained by simple aggregation of independent per-view counts. Let $C_1$ and $C_2$ denote individual-view predictions and $C_3$ the joint prediction; we report the percentage of questions for which $C_3$ agrees with either $C_1+C_2$ or $\max(C_1,C_2)$. Across models, joint predictions agree substantially more often with the maximum than with the sum (Table~\ref{tab:counting_heuristics}). In complete- and subset-overlap settings, maximum aggregation can recover the scene-level count when per-view counts are correct. In complementary views, unique instances are distributed across views, requiring information from both images. Yet maximum agreement remains consistently higher than sum agreement, despite maximum aggregation being insufficient to recover the correct scene-level count.

\begin{table*}[t]
\centering

\scriptsize
\setlength{\tabcolsep}{3pt}
\resizebox{\textwidth}{!}{%
\begin{tabular}{lccc ccc ccc}
\toprule
& \multicolumn{3}{c}{\textbf{Complete overlap}}
& \multicolumn{3}{c}{\textbf{Subset overlap}}
& \multicolumn{3}{c}{\textbf{Complementary views}} \\
\cmidrule(lr){2-4}
\cmidrule(lr){5-7}
\cmidrule(lr){8-10}

\textbf{Model}
& Acc. $\uparrow$ & Sum agr. & Max agr.
& Acc. $\uparrow$ & Sum agr. & Max agr.
& Acc. $\uparrow$ & Sum agr. & Max agr. \\
\midrule

Qwen3-VL-32B-Instruct~\cite{bai2025qwen3}
&  \textcolor{gray}{70.0} & 8.0  & 82.0
&  \textcolor{gray}{70.7} & 23.0 & 77.7
&  \textcolor{gray}{34.0} & 18.0 & 39.0 \\

InternVL3.5-38B-HF~\cite{wang2025internvl3}
&  \textcolor{gray}{72.0} & 4.0  & 77.0
& \textcolor{gray}{57.3} & 15.0 & 75.3
& \textcolor{gray}{14.5} & 11.0 & 51.0 \\

LLaVA-OneVision-2-8B-Instruct~\cite{an2026llava}
& \textcolor{gray}{55.5} & 13.5 & 68.0
& \textcolor{gray}{53.0} & 10.3   & 64.3
& \textcolor{gray}{18.0} & 4.5  & 62.0 \\

Gemma-4-31B-IT~\cite{team2026gemma}
& \textcolor{gray}{61.0} & 8.5  & 69.5
& \textcolor{gray}{63.3} & 21.7 & 70.3
& \textcolor{gray}{35.0} & 17.5 & 34.0 \\

\bottomrule

\end{tabular}%
}

\caption{
\textbf{Aggregation Heuristics.} Comparison of joint and single-view counting predictions. $C_1$ and $C_2$ denote individual-view predictions and $C_3$ the joint prediction. \textit{Acc.} is ground-truth accuracy, shown in gray for reference. \textit{Sum agr.} and \textit{Max agr.} report the percentage of questions satisfying $C_3=C_1+C_2$ and $C_3=\max(C_1,C_2)$, respectively, without using ground truth. Joint predictions often agree with the maximum in complete- and subset-overlap settings; in complementary views, maximum agreement remains higher than sum agreement despite being insufficient for scene-level counting.
}
\label{tab:counting_heuristics}

\end{table*}

\paragraph{Effect of Number of Views.}
We evaluate complete-overlap counting with $N\in\{2,3,4,5,6\}$ views while keeping the scene-level count fixed. Figure~\ref{fig:scaling} shows that performance initially improves with additional views but peaks before $N=6$ for all four models, suggesting that additional visual evidence is useful only up to the point where integrating more views becomes increasingly difficult. Moreover, there is still a substantial gap in accuracy that none of the models can close, suggesting there is plenty of room for improvement in visual reasoning capabilities when considering scenes given as multiple views.

\section{Discussion}
\label{sec:discussion}

Our results reveal a clear gap between processing multiple images and reasoning over them coherently in the current generation of VLMs. This is most evident in complementary-view counting, where performance drops substantially when models must establish cross-view correspondence and deduplicate repeated instances. Cross-view consistency also remains difficult, including for recent models, suggesting that robust scene-level multi-view understanding remains unresolved. The additional diagnostic analyses further expose limitations in how models use and integrate evidence across views.

These findings complement existing multi-view benchmarks, on which models have demonstrated stronger spatial and geometric reasoning. OneScene-Bench shows that strong spatial reasoning does not necessarily imply reliable scene-level integration. With multi-view images becoming a dominant approach for interacting with visual 3D scenes through language, OneScene-Bench offers a new perspective on what is missing for 3D reasoning through powerful LLMs.
Current models can reason about spatial relationships while still struggling with object correspondence, deduplication, and cross-view consistency, which are essential for reliable and accurate 3D reasoning.

\textbf{Limitations.} OneScene-Bench focuses on three scene-level capabilities and 1,300 manually verified examples from DL3DV-10K. While larger datasets and broader model coverage would be valuable, we prioritize careful human verification and balance evaluation breadth against computational and API costs.
Future work could further expand the benchmark across tasks, environments, and models.

\section{Conclusion}
\label{sec:conclusion}

We introduced OneScene-Bench, a controlled benchmark for evaluating whether vision-language models are able to integrate multiple views of a scene to correctly reason about visual aspects of the scene. Across open-weight and proprietary VLMs, our results revealed a persistent gap between processing multiple images and integrating them: models struggle particularly when they must establish object correspondence, reconcile complementary evidence, and detect inconsistencies across viewpoints. OneScene-Bench provides a controlled framework for measuring these capabilities and tracking progress toward more reliable scene-level multi-view understanding.

This work thus showed that models of different architectures, both closed and open, are usually unable to perform basic visual reasoning across multiple views, the prevalent representation format for 3D scenes.
Closing this gap becomes essential when moving to more complex 3D reasoning with VLMs, and our OneScene-Bench is an open benchmark designed to track progress toward this goal.

\bibliographystyle{plain}
\bibliography{references}
\newpage
\appendix

\section{Detailed Dataset Construction}
\label{app:dataset_construction}

This section provides additional details on the construction and verification
procedure used to build OneScene-Bench.

\subsection{Counting Candidate Mining}

For counting, we first randomly select a location from DL3DV-10K~\cite{ling2024dl3dv}
and randomly sample two frames from its capture sequence. We require the
difference between their frame indices to be between 10 and 25, providing
viewpoint variation while retaining sufficient visual overlap. The selected
pair is provided to Qwen3-VL-32B-Instruct~\cite{bai2025qwen3}, which is asked
to identify a clearly visible, countable, and unambiguous object category.
For each proposed category, the model estimates the number of visible
instances in each view and the number of unique physical instances across the
two views.

If no suitable object category is identified, another pair is sampled from
the same location and the process is repeated. We allow up to 10 attempts per
location; locations for which no suitable candidate is found are skipped.
This procedure is repeated across locations until sufficient candidates are
obtained for the three counting configurations.

 The final overlap type is assigned manually after verifying the visible instance sets. The resulting candidates are divided into
complete-overlap, subset-overlap, and complementary-view examples, yielding
200, 300, and 200 questions, respectively.

\subsection{Consistency Construction}
For the consistency task, we again begin by selecting two frames from the same DL3DV-10K~\cite{ling2024dl3dv} location using the 10--25 frame-index separation. To identify shared and view-specific regions, we first establish local feature correspondences between the two views using SIFT~\cite{lowe2004distinctive}. We use RANSAC~\cite{fischler1981random} to remove geometrically inconsistent matches. Regions containing a dense concentration of verified correspondences are treated as shared across the two views, while regions spatially separated from these correspondences and containing few matched features are selected as candidate view-specific regions. Image pairs for which these regions cannot be identified reliably are discarded.

We then introduce a localized object or visual artifact into one of these regions using gpt-image-1.5~\cite{openai2025gptimage15}. The model is given freedom to select an appropriate object to insert, while the target region is determined by the desired consistency condition. An edit in a view-specific region is not expected to be observed from the second viewpoint, so the resulting pair is labeled \emph{consistent}. In contrast, an edit in a shared region introduces a change that should also be observable from the second viewpoint; because it appears in only one view, the resulting pair is labeled \emph{inconsistent}.

The edited patch is resized to the target region and blended into the original image using a Gaussian-smoothed mask to reduce boundary artifacts. All generated examples are subsequently manually inspected to verify the edit, its visibility across the two viewpoints, and the resulting consistency label. Ambiguous or invalid examples are rejected or corrected. We construct 300 balanced consistency examples.

\subsection{Spatial Reasoning Candidate Mining}

For spatial reasoning, we again sample two frames from the same DL3DV-10K~\cite{ling2024dl3dv}
location with a frame-index separation of 10--25. Qwen3-VL-32B-Instruct~\cite{bai2025qwen3} is used to identify three clearly visible objects that provide a meaningful
relative-proximity comparison across the two views. The selected objects are
denoted by $A$, $B$, and $C$, and the candidate question asks whether $B$ is
physically closer to $A$ or to $C$.

A dedicated annotation tool is then used to assign consistent object names and
record the ground-truth proximity relationship. This allows the object
references and the final answer to be standardized independently of the
wording proposed during candidate generation. Candidates are retained only
when the three objects can be reliably identified across the views and the
relative-proximity relationship is unambiguous. We construct 300 spatial
reasoning questions.

\subsection{Verification Procedure}

All candidate examples are manually verified before inclusion in the
benchmark. Reviewers inspect the selected images, referenced objects, and
ground-truth annotations and reject or correct candidates with ambiguous
object identity, uncertain cross-view correspondence, unclear visibility, or
ambiguous answers.

For counting, reviewers verify the target category, visible instances in each
view, cross-view object correspondences, unique scene-level count, and
manually assigned overlap type. For consistency, reviewers verify the edited
region, whether it is shared or view-specific, and whether the resulting pair
is consistent with a single underlying scene. For spatial reasoning,
reviewers verify the three referenced objects and the relative-proximity
relationship.

The dedicated interfaces used for counting, consistency, and spatial
reasoning verification are shown in
Figures~\ref{fig:counting_sample},
\ref{fig:consistency_sample}, and
\ref{fig:spatialreasoning_sample}, respectively.

\section{Additional Details on Evaluation Tasks}
\subsection{Image Order Sensitivity}
We study whether model predictions depend on the order in which the two views are presented. A model that forms a consistent scene representation should produce similar predictions when the two views are swapped. We evaluate four representative open-weight models on the same questions using both the original and reversed image orders, and measure the absolute change in accuracy across all five evaluation settings. Most model--task combinations are stable, with 18 of 20 changing by at most four percentage points. The largest changes occur for Gemma-4~\cite{team2026gemma} on complete-overlap counting and LLaVA-OneVision-2-8B-Instruct~\cite{an2026llava} on spatial reasoning, with differences of 6.0 and 9.0 points, respectively. Overall, models are reasonably robust to image ordering, suggesting that input order is not a major source of variation in the benchmark.

\begin{table}[t]
\centering
\captionsetup{position=bottom,skip=7pt}
\resizebox{\columnwidth}{!}{
\begin{tabular}{lccccc}
\toprule
& \multicolumn{3}{c}{\textbf{Counting}}
& \textbf{Consistency $\downarrow$}
& \textbf{Spatial} \\
\cmidrule(lr){2-4}
\textbf{Model}
& \textbf{Complete $\downarrow$}
& \textbf{Subset $\downarrow$}
& \textbf{Complementary $\downarrow$}
&
& \textbf{Reasoning $\downarrow$}\\
\midrule
Qwen3-VL-32B-Instruct~\cite{bai2025qwen3}         & 0.5 & 2.4 & 1.5 & 4.0 & 2.7 \\
Gemma-4-31B-IT~\cite{team2026gemma}                & 6.0 & 1.6 & 4.0 & 0.3 & 0.3 \\
InternVL3.5-38B-HF~\cite{wang2025internvl3}           & 3.5 & 1.0 & 1.5 & 2.3 & 1.0 \\
LLaVA-OneVision-2-8B-Instruct~\cite{an2026llava} & 1.5 & 0.3 & 0.5 & 0.0 & 9.0 \\
\bottomrule
\end{tabular}
}

\caption{
\textbf{Image-Order Sensitivity.}
Absolute difference in accuracy (percentage points) between the original and reversed input-image orders, $\lvert\mathrm{Acc}_{\mathrm{original}}-\mathrm{Acc}_{\mathrm{reverse}}\rvert$. Overall, 18 of 20 model--task combinations change by at most four points, indicating broad robustness to image order, with larger sensitivity for Gemma-4 on complete-overlap counting and LLaVA-OneVision-2 on spatial reasoning.
}
\label{tab:image_order}
\end{table}

\subsection{No-Image Evaluation Setting}
\label{app:refusal}

To further characterize model behavior when visual evidence is unavailable, we
record whether models explicitly refuse to answer in the no-image setting.
A response is counted as a refusal only when the model explicitly states that
the question cannot be answered without the missing visual information;
incorrect or guessed answers are not counted as refusals. As shown in
Table~\ref{tab:no_image_refusals}, Qwen3-VL-32B-Instruct~\cite{bai2025qwen3}, InternVL3.5-38B-HF~\cite{wang2025internvl3},
and LLaVA-OneVision-2-8B-Instruct~\cite{an2026llava} do not refuse any questions. In contrast,
Gemma-4-31B-IT~\cite{team2026gemma} refuses all consistency questions and 293 of 300 spatial
reasoning questions, while refusing only 21 of 700 counting questions. This
difference shows that models vary substantially in whether they recognize
when visual information is necessary to answer a question.

\begin{table*}[t]
\centering
\small
\setlength{\tabcolsep}{5pt}

\resizebox{\textwidth}{!}{%
\begin{tabular}{lrrrrr}
\toprule
& \multicolumn{3}{c}{\textbf{Counting}}
& \textbf{Consistency}
& \textbf{Spatial} \\
\cmidrule(lr){2-4}

\textbf{Model}
& \textbf{Complete}
& \textbf{Subset}
& \textbf{Complementary}
&
& \textbf{Reasoning}\\

& $(n=200)$
& $(n=300)$
& $(n=200)$
& $(n=300)$
& $(n=300)$ \\
\midrule

Qwen3-VL-32B-Instruct~\cite{bai2025qwen3}
& 0 & 0 & 0 & 0 & 0 \\

InternVL3.5-38B-HF~\cite{wang2025internvl3}
& 0 & 0 & 0 & 0 & 0 \\

LLaVA-OneVision-2-8B-Instruct~\cite{an2026llava}
& 0 & 0 & 0 & 0 & 0 \\

Gemma-4-31B-IT~\cite{team2026gemma}
& 6 & 3 & 12 & 300 & 293 \\

\bottomrule
\end{tabular}%
}

\caption{
\textbf{No-image refusal count.}
Number of explicit refusals in the no-image setting. A refusal is counted
only when a model explicitly states that the question cannot be answered
without visual input; incorrect or guessed answers are not counted.
}
\label{tab:no_image_refusals}
\end{table*}

\section{Evaluation Prompts}
\label{app:prompts}

For reproducibility, we provide the exact prompts used for model evaluation. The same task prompts are used for all evaluated models. The free-form model responses are evaluated using an LLM judge, which receives the raw model response and the human-verified ground-truth answer.

\subsection{Task Prompts}

\paragraph{Multi-View Counting.}
\par\noindent
\begin{promptbox}
These two images show the same scene from different viewpoints.
How many distinct instances of '\{object\_name\}' are visible in the scene?
Do not double-count the same instance if it appears in both views.
Use both images jointly to infer the scene-level count.
Reply with only one integer and no explanation.
\end{promptbox}

Here, \texttt{\{object\_name\}} is replaced by the target object category for each example. The same prompt is used for all three counting configurations. For the multi-image ablation, we slightly modify the counting prompt to refer to multiple input images rather than two images, while keeping the remaining instructions unchanged.

\paragraph{Cross-View Consistency.}
\par\noindent
\begin{promptbox}
These two images are different views of a scene.
Check whether they are visually consistent.

Answer YES if the images show the same scene contents without any clear contradiction.
Answer NO if there is any clear inconsistency, such as a missing object, extra object, changed object, or contradictory scene content.

Write exactly:
Answer: YES
or
Answer: NO
\end{promptbox}

\paragraph{Multi-View Spatial Reasoning.}
\par\noindent
\begin{promptbox}
These two images show the same scene from different viewpoints.

A = {a}
B = {b}
C = {c}

Task:
Decide whether B is closer to A or C.

Output exactly:
Answer: <A or C>
\end{promptbox}

Here, \texttt{\{a\}}, \texttt{\{b\}}, and \texttt{\{c\}} are replaced by the corresponding object names for each example.

\subsection{LLM-as-a-Judge Prompts}

\paragraph{Counting Judge.}
\par\noindent
\begin{promptbox}
You are evaluating the answer of a vision-language model on a counting task.

The model was asked to count the number of distinct instances of an object across multiple views of the same scene.

Object:
'\{object\_name\}'

Human-verified ground-truth count:
'\{ground\_truth\}'

Model's raw response:
'\{raw\_output\}'

Determine whether the model's response gives the same final count as the human-verified ground truth.

Accept semantically equivalent representations of the number. For example:
- "3"
- "three"
- "There are three."
- "The answer is 3."
should all be treated as the count 3.

Ignore irrelevant wording or explanation as long as the model's final intended count is unambiguous.

If the model gives multiple conflicting counts and no unambiguous final answer, mark it INCORRECT.

Output exactly one word:
CORRECT
or
INCORRECT
\end{promptbox}
\noindent
Here, \texttt{\{object\_name\}} denotes the target object category,
\texttt{\{ground\_truth\}} is the human-verified scene-level count, and
\texttt{\{raw\_output\}} is the complete response generated by the evaluated model.

\paragraph{Consistency Judge.}
\par\noindent
\begin{promptbox}
You are evaluating the answer of a vision-language model on a binary visual consistency task.

The original task asked the model to determine whether two views of a scene are visually consistent.

The human-verified ground-truth answer is:
'\{ground\_truth\}'
The model's raw response is:
'\{raw\_output\}'

Determine whether the model's final intended answer agrees with the human-verified ground truth.

Interpret the following as equivalent:
YES = SAME = CONSISTENT
NO = DIFFERENT = INCONSISTENT

The model may include reasoning, explanation, or additional wording.
Ignore such wording as long as the final intended answer is unambiguous.

For example:
Ground truth: YES
Model response: "The two images appear to show the same scene.
Answer: YES"
=> CORRECT
Ground truth: NO
Model response: "There is an object missing in the second image, so the scenes are inconsistent."
=> CORRECT
Ground truth: NO
Model response: "The images are consistent. Answer: YES"
=> INCORRECT

If the model gives conflicting answers and there is no unambiguous final answer, mark it INCORRECT.

Output exactly one word:
CORRECT
or
INCORRECT
\end{promptbox}
\noindent
Here, \texttt{\{ground\_truth\}} is the human-verified consistency label
(\texttt{YES} or \texttt{NO}), and \texttt{\{raw\_output\}} is the complete
response generated by the evaluated model.

\paragraph{Spatial Reasoning Judge.}
\par\noindent
\begin{promptbox}
You are evaluating the answer of a vision-language model on a spatial reasoning task.

The original task asked the model to determine whether object B is closer to object A or object C.

The objects were:

A = '\{object\_a\}'
B = '\{object\_b\}'
C = '\{object\_c\}'

The human-verified ground-truth answer is: '\{ground\_truth\}'

The model's raw response is: '\{raw\_output\}'

Determine whether the model's final intended answer agrees with the human-verified ground truth.

The answer must correspond to either A or C.
Accept semantically equivalent responses.

Examples that mean A:
- "A"
- "Answer: A"
- "B is closer to A."
- "Object A."
- "The '\{object\_b\}' is closer to the '\{object\_a\}'."

Examples that mean C:
- "C"
- "Answer: C"
- "B is closer to C."
- "Object C."
- "The '\{object\_b\}' is closer to the '\{object\_c\}'."

The model may include reasoning, explanation, or additional wording. Ignore such wording as long as the final intended answer is unambiguous.

If the model gives multiple conflicting answers and there is no unambiguous final answer, mark it INCORRECT.

Output exactly one word:
CORRECT
or
INCORRECT
\end{promptbox}
\noindent
Here, \texttt{\{object\_a\}}, \texttt{\{object\_b\}}, and
\texttt{\{object\_c\}} denote the three referenced objects,
\texttt{\{ground\_truth\}} is the human-verified answer
(\texttt{A} or \texttt{C}), and \texttt{\{raw\_output\}} is the complete
response generated by the evaluated model.

\clearpage
\section{Screenshots from the Dataset Verification Tool}
\begin{figure}[h]
    \centering
    \includegraphics[width=0.9\linewidth]{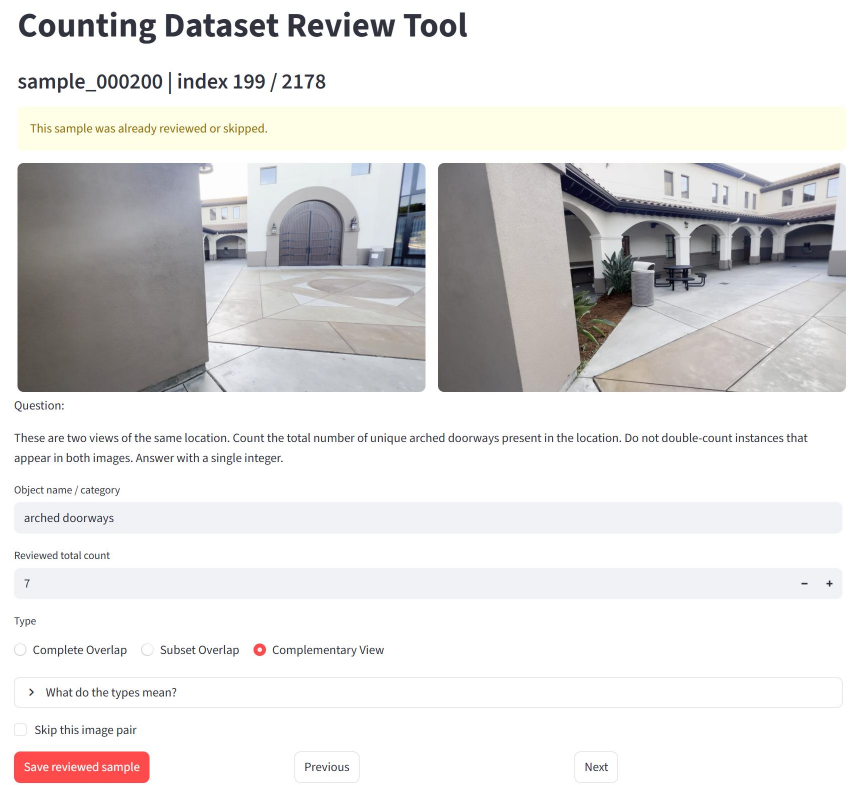}
    \caption{\textbf{Screenshot of the manual counting review interface}. The interface displays the selected multi-view image pair, target object category, reviewed scene-level count, and manually assigned overlap type. Reviewers can also skip ambiguous image pairs before finalizing the
annotation.}
    \label{fig:counting_sample}
\end{figure}

\begin{figure}[t]
    \centering
    \includegraphics[width=0.9\linewidth]{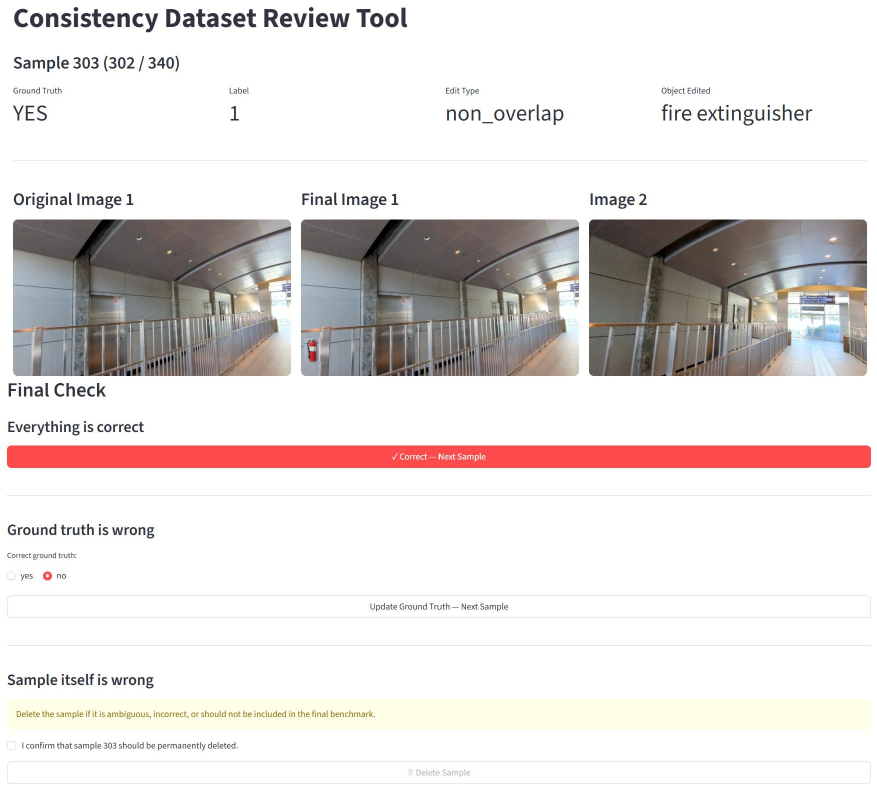}
    \caption{\textbf{Screenshot of the consistency review tool}. The interface displays the original and edited views,
ground-truth label, edit type, and edited object. Reviewers can confirm the
example, correct an incorrect ground-truth label, or permanently remove an
invalid or ambiguous example.}
    \label{fig:consistency_sample}
\end{figure}

\begin{figure}[t]
    \centering
    \includegraphics[width=0.9\linewidth]{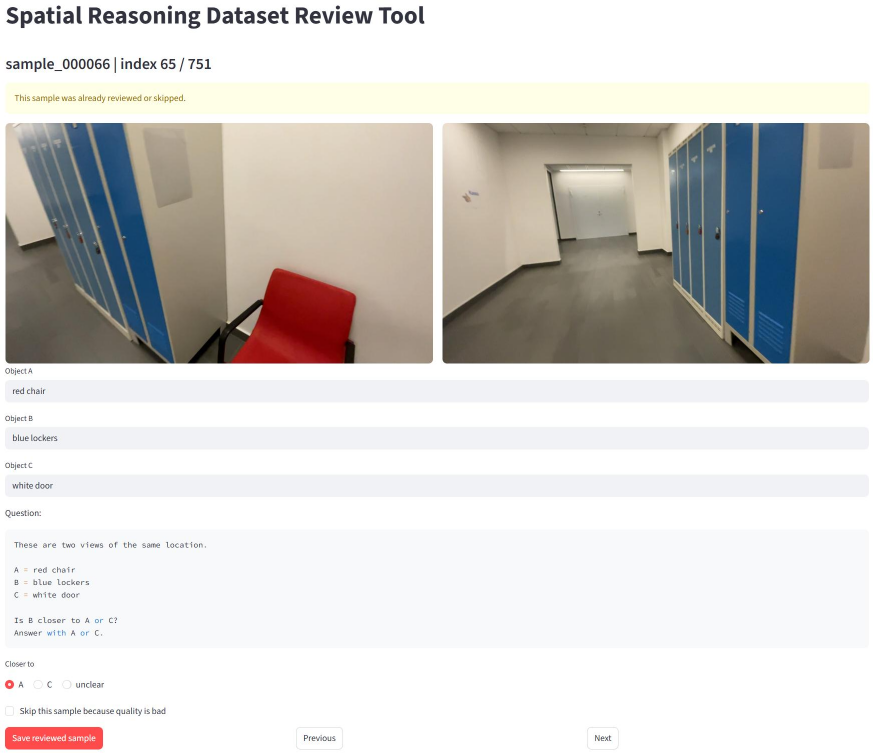}
    \caption{\textbf{Screenshot of the spatial-reasoning verification interface}. The tool displays the two views, the three referenced objects ($A$, $B$, and $C$), the generated
question, and the reviewer controls for assigning the ground-truth proximity
relationship or marking the example for removal.}
    \label{fig:spatialreasoning_sample}
\end{figure}

Figures~\ref{fig:counting_sample}, \ref{fig:consistency_sample}, and \ref{fig:spatialreasoning_sample} show the interfaces used during manual verification of OneScene-Bench examples for the three benchmark tasks.

\clearpage
\section{Qualitative Samples}
\begin{figure}[h]
    \centering
    \includegraphics[width=0.99\linewidth]{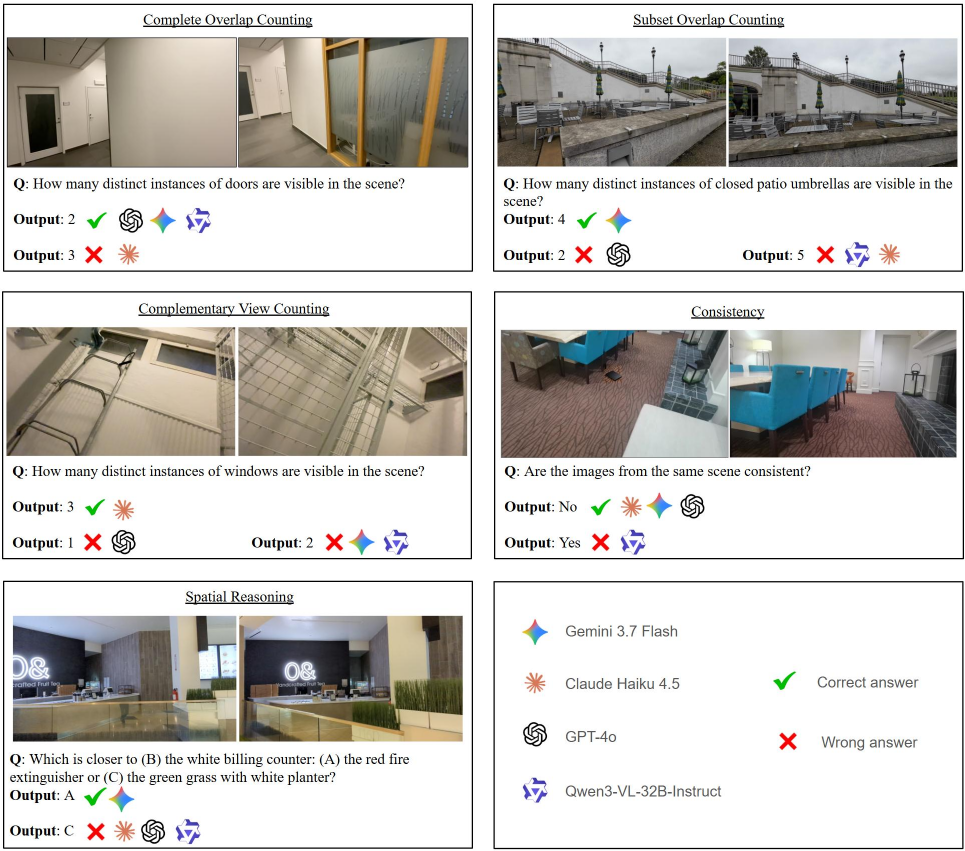}
    \caption{
\textbf{Representative examples from the five tasks in OneScene-Bench}: multi-view
counting under complete-overlap, subset-overlap, and complementary-view
configurations, cross-view consistency, and multi-view spatial reasoning.
Questions are slightly shortened for visual presentation while preserving their
original meaning. Model predictions are shown using the corresponding model
logos, with green checkmarks and red crosses indicating correct and incorrect
predictions, respectively.
}
    \label{fig:qual_samples}
\end{figure}

Figure~\ref{fig:qual_samples} shows example benchmark questions from all five tasks, along with the answers provided by four different VLMs. The ground-truth answer is marked with a green tick.

\section{Compute Resources}
\label{app:resources}
Open-source models were evaluated primarily on NVIDIA A100 GPUs with 80 GB of GPU memory. The closed-source models were evaluated through their respective paid APIs. The evaluation of an open-source model required approximately 2 hours per task.

\section{Licenses for Existing Assets}
\label{license}
OneScene-Bench is constructed from DL3DV-10K~\cite{ling2024dl3dv}, which is used in accordance with the DL3DV-10K Terms of Use. The open-weight models used in our experiments are accessed and used under their respective licenses, including Apache License 2.0 for Qwen3-VL~\cite{bai2025qwen3}, InternVL3.5~\cite{wang2025internvl3}, LLaVA-OneVision~\cite{li2024llava, an2026llava}, and Gemma 4~\cite{team2026gemma}, and the MIT License for Kimi-VL~\cite{team2025kimi} and GLM-4.6V-Flash~\cite{hong2025glm}. Proprietary models are accessed through their official APIs in accordance with the applicable terms of service.

\newpage

\end{document}